\pdfoutput=1

\documentclass[11pt]{article}

\usepackage{acl}
\usepackage[colorinlistoftodos]{todonotes}
\usepackage{times}
\usepackage{latexsym}
\usepackage{multirow}
\usepackage{float}
\usepackage{placeins}

\usepackage[T1]{fontenc}

\usepackage[utf8]{inputenc}

\usepackage{microtype}

\title{Transformers on Multilingual Clause-Level Morphology}

\author{Emre Can Acikgoz \\
  KUIS AI, Koç University \\
  \texttt{eacikgoz17@ku.edu.tr} \\\And
  Tilek Chubakov \\
  KUIS AI, Koç University\\
  \texttt{tchubakov@ku.edu.tr} \\\And
  Müge Kural \\
  KUIS AI, Koç University\\
  \texttt{mugekural@ku.edu.tr} \\\AND
  Gözde Gül Şahin \\
  KUIS AI, Koç University\\
  \texttt{gosahin@ku.edu.tr} \\\And
  Deniz Yuret \\
  KUIS AI, Koç University\\
  \texttt{dyuret@ku.edu.tr} \\}

\begin{document}
\maketitle
\begin{abstract}
This paper describes our winning systems in MRL: The 1st Shared Task on Multilingual Clause-level Morphology (EMNLP 2022 Workshop) designed by KUIS AI NLP team. We present our work for all three parts of the shared task: inflection, reinflection, and analysis. We mainly explore transformers with two approaches: (i) training models from scratch in combination with data augmentation, and (ii) transfer learning with prefix-tuning at multilingual morphological tasks. Data augmentation significantly improves performance for most languages in the inflection and reinflection tasks. On the other hand, Prefix-tuning on a pre-trained mGPT model helps us to adapt analysis tasks in low-data and multilingual settings. While transformer architectures with data augmentation achieved the most promising results for inflection and reinflection tasks, prefix-tuning on mGPT received the highest results for the analysis task. Our systems received 1st place in all three tasks in MRL 2022.\footnote{\url{https://github.com/emrecanacikgoz/mrl2022}}
\end{abstract}

\section{Introduction}

The shared task on multilingual clause-level morphology was designed to provide a benchmark for morphological analysis and generation at the level of clauses for various typologically diverse languages. The shared task is composed of three subtasks: \textit{inflection}, \textit{reinflection} and \textit{analysis}. For the inflection task, participants are required to generate an output clause, given a verbal lemma and a specific set of morphological tags (features) as an input. In the reinflection task the input is an inflected clause, accompanied by its features (tags). Participants need to predict the target word given a new set of tags (features). Finally, the analysis task requires predicting the underlying lemma and tags (features) given the clauses.  

Literature has examined morphology mainly at the word level, but morphological processes are not confined to words. Phonetic, syntactic, or semantic relations can be studied at phrase-level to explain these processes. Thus, this shared task examines phrase-level morphology and questions the generalization of the relations between the layers of language among languages with different morphological features. The shared task includes eight languages with different complexity and varying morphological characteristics: English, French, German, Hebrew, Russian, Spanish, Swahili, and Turkish. 

\begin{table}[]
\centering
\resizebox{\columnwidth}{!}{%
\begin{tabular}{|l|l|l|}
\hline
\multicolumn{3}{|c|}{\textbf{Task1: Inflection}}                                                                                         \\\hline
\multirow{2}{*}{Source} & Lemma            & give                                                                                      \\
                        & Features         & \begin{tabular}[c]{@{}l@{}} IND;FUT;NOM(1,SG);\\ACC(3,SG,MASC);DAT(3,SG,FEM)\end{tabular}   \\\hline
Target                  & Clause           & I will give him to her                                                                    \\[0.5ex]\hline
\multicolumn{3}{|c|}{\textbf{Task2: Reinflection}}                                                                                       \\\hline
\multirow{3}{*}{Source} & Clause           & I will give him to her                                                                    \\
                        & Features         & \begin{tabular}[c]{@{}l@{}} IND;FUT;NOM(1,SG);\\ ACC(3,SG,MASC);DAT(3,SG,FEM)\end{tabular} \\
                        & Desired Features & \begin{tabular}[c]{@{}l@{}} IND;PRS;NOM(1,PL);\\ ACC(2);DAT(3,PL);NEG\end{tabular}         \\\hline
Target                  & Desired Clause   & We don't give you to them                                                                 \\[0.5ex]\hline
\multicolumn{3}{|c|}{\textbf{Task3: Analysis}}                                                                                           \\\hline
Source                  & Clause           & I will give him to her                                                                    \\\hline
\multirow{2}{*}{Target} & Lemma            & give                                                                                      \\
                        & Features         & \begin{tabular}[c]{@{}l@{}} IND;FUT;NOM(1,SG);\\ ACC(3,SG,MASC);DAT(3,SG,FEM)\end{tabular} \\\hline
\end{tabular}
}
\caption{Description of the each three task: inflection, reinflection, analysis. \textbf{Task1 (Inflection).} For the given lemma and the features, target is the desired clause.\textbf{Task2 (Reinflection).} Input is the clause, its features, and the desired output features. Target is the desired clause that represented by the desired features in the source. \textbf{Task3 (Analysis).} For a given clause, output is the corresponding lemma and the morphological features.}
\label{tab:tasks}
\end{table}

In our work, we explored two main approaches: (1) training character-based transformer architectures from scratch with data augmentation, (2) adapting a recent prefix-tuning method for language models at multilingual morphological tasks.


\section{Methods}
In this section, first we cover the model architectures and training strategies that we have used \cite{Vaswani2017, mgpt:22, prefixtunning:21}, and then discuss our data augmentation strategies in details \cite{hall:19}.

\subsection{Vanilla Transformer}
\label{section: transformer}

We used a modified version of vanilla Transformer architecture in \citet{Vaswani2017} which contains 4 layers of encoder and decoder with 4 multi-head attentions. The embedding size and the feed-forward dimension is set to 256 and 1024, respectively. As suggested in \citet{tagtransformer:21}, we used layer normalization before the self-attention and feed-forward layers of the network that leads to slightly better results. We used these in inflection and reinflections tasks.

\subsection{Prefix-Tuning}
\label{section: prefix}
\begin{figure}
    \centerline{\includegraphics[width=0.5\textwidth]{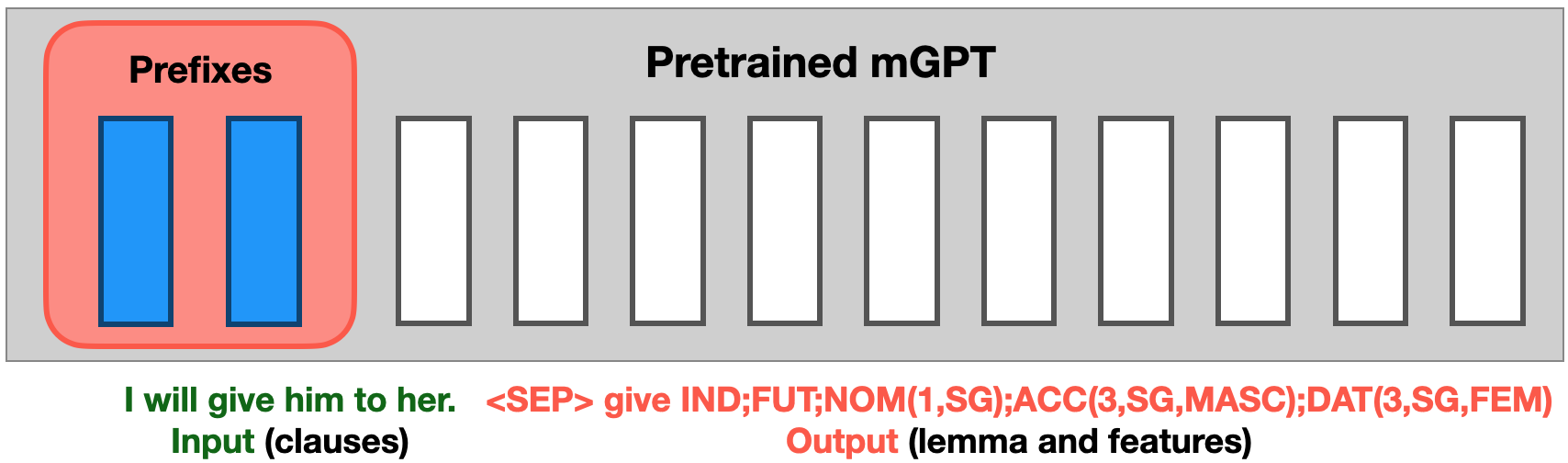}}
    \caption{\textbf{Task3 (Analysis)} example by using prefix-tuning method. We freeze all the parameters of the pre-trained mGPT model and only optimize the prefix, which are shown inside the red block. Each vertical block denote transformer activations at one time step.}
    \label{fig:prefix1}
\end{figure}

Using prefix-tuning reduces computational costs by optimizing a small continuous task-specific vectors, called prefixes, while keeping frozen all the other parameters of the LLM. We added two prefixes, called virtual tokens in \citet{prefixtunning:21}, the gradient optimization made across these prefixes that is described in the Figure \ref{fig:prefix1}. We used \citet{mgpt:22} weights during prompting. Prefix-tuning method outperforms other fine-tuning approaches in low-data resources and better adapts to unseen topics during prompting \cite{prefixtunning:21}.

\subsection{Data Augmentation}
\label{section: augment}
\begin{figure}
    \centerline{\includegraphics[width=0.5\textwidth]{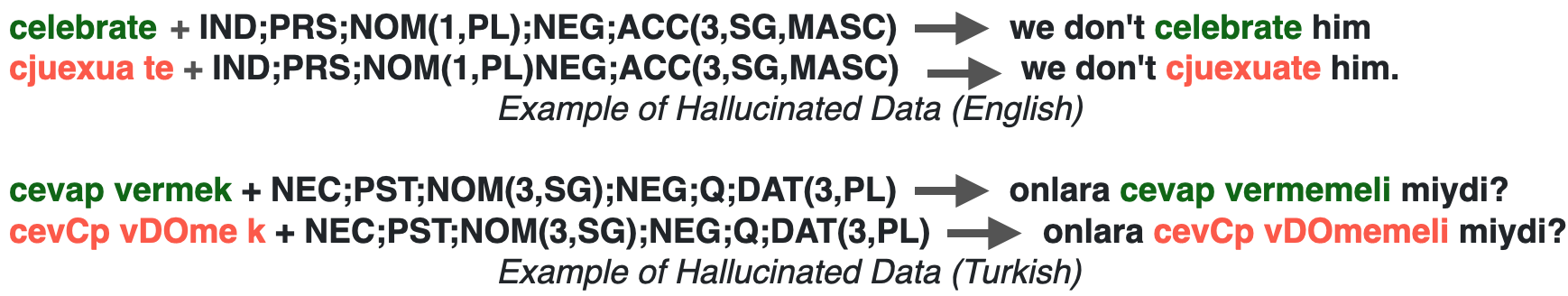}}
    \caption{In order to create the hallucinated samples, we first align the characters of the lemma and the inflected forms. After that, we substitute the stem parts of the input with random characters that comes from the validation set and test set, as shown in the figure.}
    \label{fig:hall}
\end{figure}

Hallucinating the data for low-resource languages results with a remarkable performance increase for inflection \citet{hall:19}. The hallucinated data is generated by replacing the stem characters of the aligned word with random characters by using the validation or test sets (see Fig. \ref{fig:hall}). This way, the amount increase in the training data helps the model to learn and generalize rare seen samples. On the other hand, the amount of hallucinated data that will be added to the training set, hyperparameter $N$, is also another parameter that directly effects our accuracy. Therefore, hyperparameter $N$ needs to be decided specifically for each language according to corresponding language's complexity and topology.

\section{Experimental Settings}

\subsection{Dataset}

In the shared task, there are eight different languages with varying linguistic complexity which comes from different language families: English, French, German, Hebrew, Russian, Swahili, Spanish, Turkish. For Hebrew there are two versions as Hebrew-vocalized and Hebrew-unvocalized. Training data contains 10,000 instances for each language and there are 1,000 samples both in development set and test set. Swahili and Spanish are the surprise languages that announced two weeks before the final submission day, together with the unlabeled test data for each language.  

\subsection{Evaluation}

Models are evaluated according to Exact Match (EM), Edit Distance (ED), and F1 accuracy. For task1 (inflection) and task2 (reinflection) ED is the leaderboard metric. For task3 (analysis), F1 score is the objective. EM accuracy represents the ratio of correctly predicted lemma and features, and ED is calculated based on Levenshtein Distance which indicates how different two strings are, (the ground truth and prediction for our case) from each other. F1 accuracy is the harmonic mean of the precision and recall. F1 accuracy is upweighted for the lemma score in our task. In the leaderboard, the results are averaged across each language.

\FloatBarrier
\begin{table*}[h]
\centering
\begin{tabular}{lccc|ccc|ccc} \\
 & \multicolumn{3}{c}{\textbf{Task1: Inflection}}  & \multicolumn{3}{c}{\textbf{Task2: Reinflection}}             & \multicolumn{3}{c}{\textbf{Task3: Analysis}} \\ \hline
\textbf{Model}     & \multicolumn{3}{c}{\textbf{Transformer + D.A.}} & \multicolumn{3}{|c|}{\textbf{Transformer}} & \multicolumn{3}{|c}{\textbf{Prefix Tuning}}   \\ \hline
\textbf{Metrics}   & F1$\uparrow$  & EM$\uparrow$ & ED$\downarrow$   & F1$\uparrow$ & EM$\uparrow$ & ED$\downarrow$      & F1$\uparrow$    & EM$\uparrow$    & ED$\downarrow$    \\ \hline
Deu                & 97.71 & 91.80 & 0.241                           & 92.40 & 66.50 & 0.788                             & 95.89 & 83.40 & 0.991 \\
Eng                & 98.02 & 88.90 & 0.221                           & 95.42 & 72.30 & 0.477                             & 99.61 & 98.50 & 0.064 \\
Fra                & 98.59 & 93.20 & 0.124                           & 92.64 & 68.30 & 0.758                             & 95.63 & 81.90 & 0.933 \\
Heb                & 97.73 & 89.80 & 0.550                           & 94.00 & 83.30 & 0.796                             & 92.84 & 73.50 & 1.322 \\
Heb-Unvoc          & 97.96 & 94.20 & 0.113                           & 86.70 & 57.70 & 1.002                             & 82.09 & 36.20 & 2.044 \\
Rus                & 97.57 & 87.70 & 0.828                           & 97.29 & 84.90 & 0.854                             & 97.51 & 88.60 & 3.252 \\
Swa                & 99.72 & 99.61 & 0.019                           & 92.05 & 84.47 & 0.182                             & 90.51 & 62.63 & 3.114 \\
Spa                & 98.79 & 92.00 & 0.199                           & 96.42 & 77.60 & 0.480                             & 98.11 & 89.40 & 0.560 \\
Tur                & 97.50 & 89.80 & 0.333                           & 95.36 & 84.70 & 0.593                             & 95.36 & 84.70 & 0.593 \\\hline
Average            & 91.89 & 98.18 & \textbf{0.292}                  & 93.14 & 74.72 & \textbf{0.705}                    & \textbf{94.17} & 77.65 & 1.430 \\ \hline
\end{tabular}
\caption{Results on the test sets for all tasks and languages with the corresponding models. Edit Distance is the leaderboard ranking metric for Task1: Inflection and Task2: Reinflection, and F1 score is used for leaderboard ranking in Task3: Analysis. D.A. indicates data augmentation.}
\label{tab:results-table}
\end{table*}
\FloatBarrier

\subsection{Shared Task}
Multilingual Clause-level Morphology (MRL 2022) contains three different tasks as Task1: Inflection, Task2: Reinflection, and Task3: Analysis. As KUIS AI team, we have attended each of them separately.

\subsubsection{Task1: Inflection}
The goal of the task is to produce the output clause and its features forgiven verbal lemma and a set of morphological features, see Table \ref{tab:tasks}. For inflection task, we have trained a vanilla transformer model from scratch by adding some hallucinated data for the training set. The data hallucination method, discussed in \ref{section: augment}, improved our results significantly. As suggested in \citet{tagtransformer:21}, we observed the effect of the large batch sizes that results with an increase in accuracy. Thus, we set the batch size to 400 and we trained our model for 20 epochs. We used Adam optimizer by setting $\beta_1$ to 0.9 and $\beta_2$ to 0.98. We started with a learning rate of 0.001 with 4,000 warm-up steps. Then, we decrease it with the inverse of the square-root for the remaining steps. We have used label smoothing with a factor of 0.1 and applied the same dropout rate of 0.3.

\subsubsection{Task2: Reinflection}
In reinflection the task is to generate the desired output format as in inflection; however, the input is consist of an inflected clause, its corresponding features, and a new set of features that represents the desired output form. We again use the same vanilla Transformer architecture, and exactly the same training parameters that we have used in inflection task. We tried both (i) giving the all source data as input, and (ii) using only the inflected clause and its desired features. We have examined that, both our EM and ED accuracy increased in a large manner when we ignore source clause's features in input before feeding it to the model. 

\subsubsection{Task3: Analysis}
Analysis task can be seen as the opposite of the inflection task. For given clauses and its features, we try to generate the lemma and the corresponding morphological features. We used the prefix-tuning method for the analysis task. The prefix template was given as the source and the features were masked. During prompting, we gave the clause-level in input and the target lemma together with its features were expected from the output, like a machine translation task. The source and target are given together with the trainable prefixes, i.e. continuous prompt vectors, and the gradient optimization made across these prefixes. For the mGPT-based Prefix-Tuning model, we have used the \textit{Huggingface}, \citet{huggingface} and the corresponding model weights \textit{sberbank-ai/mGPT}. The prefixes were trained for 10 epochs with a batch size of 5 due computational resource constraints. We used Adam optimizer with weight decay fix which is introduced in \citet{adamw} with $\beta_1$=0.9 and  $\beta_2$=0.999. The learning rate is initialized to $5\times10^{-5}$ and a linear scheduler is used without any warm-up steps.

\subsection{Results}
Our submitted results are provided in Table \ref{tab:results-table}. The announced results by the shared task are in the Table \ref{tab:mrl_results} which are evaluated among the provided unlabeled test set.

For the inflection task, with the help of data augmentation, we have achieved best average edit distance for languages. Specially, for Swahili the edit distance is nearly perfect as well as the exact match. It is followed by Hebrew-Unvoc and French. We observed the highest edit distance and the lowest exact match scores for Russian. At the end, we observed that, reducing edit distance does not always bring better exact match.

For the reinflection task, using trained transformer models from scratch, we again see the best results for Swahili with the lowest edit distance. This time, the highest edit distance belongs to Hebrew-Unvoc as well as the lowest exact match. The number of words and characters in the examples of task datasets may be the factors and should also be considered.

Finally for the analysis, with the help of prefix-tuning, we achieved the best results for English with highest F1 score. The ease of finding English pre-trained models led us to experiment with English-only GPT models, and we subsequently discovered that multilingual GPT gives better results when using prefix-tuning. Tuning on mGPT has the lowest performance with Hebrew-Unvoc, due the low ratio of training samples in Hebrew during pre-training compared to other languages.

\begin{table}[]
\centering
\resizebox{\columnwidth}{!}{%
\begin{tabular}{lccc}
\hline
\\[-1em]
\textbf{System}      & \textbf{Inflection}      &\textbf{ Reinflection}      & \textbf{Analysis}        \\\hline\\[-1em]
Transformer Baseline & 3.278                    & 4.642                      & 80.00                    \\ \\[-1em]
mT5 Baseline         & 2.577                    & 2.826                      & 84.50                    \\\\[-1em]
KUIS AI              & \textbf{0.292}           & \textbf{0.705}             & \textbf{94.17}           \\ \\[-1em]\hline
\end{tabular}%
}
\caption{Submitted results for MRL shared task that is averaged across 9 languages. Metrics for the inflection and reinflection tasks is the edit distance, and for analysis the metric is averaged F1 score with the lemma being treated as an up-weighted feature.}
\label{tab:mrl_results}
\end{table}

\section{Related Work}

Word-level morphological tasks have been studied to a great extent, with LSTM \cite{Wu2019ExactHM,Cotterell2016TheS2,malaviya-etal-2019-simple, DBLP:conf/acl/SteedmanS18}, GRU \cite{conforti-etal-2018-neural}, variants of Transformer \citet{Vaswani2017,tagtransformer:21} and other neural models (e.g., invertible neural networks~\cite{SahinG20}). Unlike word-level, there is limited work on clause-level morpho-syntactic modeling. \citet{Goldman2022MorphologyWB} presents a new dataset for clause-level morphology covering 4 typologically-different languages (English, German, Turkish, and Hebrew); motivates redefining the problem at the clause-level to enable the cross-linguistical study of neural morphological modeling; and derives clause-level inflection, reinflection, and analysis tasks together with baseline model results. 

Pre-trained LLMs have been successfully applied to downstream tasks like sentiment analysis, question answering, named entity recognition, and part-of-speech~(POS) tagging \cite{devlin-etal-2019-bert, Yang2019XLNetGA, JMLR:v21:20-074}. Even though, there is limited work on applications of LLMs to morphological tasks, it has been demonstrated that using pre-trained contextualized word embeddings can significantly improve the performance of models for downstream morphological tasks. \citet{Inoue2022MorphosyntacticTW} explored BERT-based classifiers for training morphosyntactic tagging models for Arabic and its dialect. \citet{Anastasyev2020EXPLORINGPM} explored the usage of ELMo and BERT embeddings to improve the performance of joint morpho-syntactic parser for Russian. \citet{Hofmann2020DagoBERTGD} used a fine-tuning approach to BERT for the derivational morphology generation task. Finally, \citet{2021AlephBERTPA} presented a large pre-trained language model for Modern Hebrew that shows promising results at several tasks.

On the other hand, since fine-tuning LLMs requires to modify and store all the parameters in a LM that results with a huge computational cost. \citet{Rebuffi:17, Houlsby:19} used adapter-tuning which adds task-specific layers (adapters) between the each layer of a pre-trained language model and tunes only the 2\%-4\% parameters of a LM. Similarly, \citet{prefixtunning:21} proposed prefix-tuning which is a light-weight alternative method for adapter-tuning that is inspired by prompting.

\section{Conclusion}
In this paper, we described our winning methods multilingual clause-level morphology shared task for inflection, reinflection, and analysis. Due to the different complexity between tasks and the varying morphological characteristics of languages, there is no single best model that achieves the best results for each task in each language. Thus, we try to implement different types of systems with different objectives. For inflection we used a vanilla Transformer adapted from \citet{Vaswani2017} and applying data hallucination substantially improves accuracy \cite{hall:19}. The reinflection task is more challenging compared to the other tasks due to its complex input form. To overcome this issue, we have removed the original feature tags from the input. We only used the inflected clause and target features in the input. We again used a vanilla Transformer as a model choice. Finally, for the analysis task, we used the prefix-tuning method based on mGPT. On average, we have achieved the best results for every three tasks among all participants.

\section*{Acknowledgements}

This work is supported by KUIS AI Center from Koç University, Istanbul. We gratefully acknowledge this support. Last but not least, we would like to kindly thank our organizers for answering our questions and for the effort they have made to fix the issues that we struggled during the competition process.
\bibliography{anthology,custom}
\bibliographystyle{acl_natbib}




\end{document}